\documentclass[runningheads]{llncs}
\usepackage{graphicx}
\usepackage{comment}
\usepackage{amsmath,amssymb}
\usepackage{color}
\usepackage{tabularx} 

\usepackage[width=122mm,left=12mm,paperwidth=146mm,height=193mm,top=12mm,paperheight=217mm]{geometry}

\usepackage{hyperref}

\usepackage{makecell}

\usepackage{enumitem}

\newlist{SubItemList}{itemize}{1}
\setlist[SubItemList]{label={$-$}}

\let\OldItem\item
\newcommand{\SubItemStart}[1]{%
    \let\item\SubItemEnd
    \begin{SubItemList}[resume]%
        \OldItem #1%
}
\newcommand{\SubItemMiddle}[1]{%
    \OldItem #1%
}
\newcommand{\SubItemEnd}[1]{%
    \end{SubItemList}%
    \let\item\OldItem
    \item #1%
}
\newcommand*{\SubItem}[1]{%
    \let\SubItem\SubItemMiddle%
    \SubItemStart{#1}%
}%

\begin{document}
\pagestyle{headings}
\mainmatter
\def\ECCVSubNumber{27}  

\title{CA-GAN: Weakly Supervised Color Aware GAN for Controllable Makeup Transfer} 

\author{Robin Kips\inst{1,2}\and
Pietro Gori\inst{2}\and
Matthieu Perrot\inst{1} \and
Isabelle Bloch \inst{2}}
\authorrunning{R. Kips et al.}
\titlerunning{CA-GAN: Weakly Supervised Color Aware GAN}
%
\institute{L'Oréal Research and Innovation, France \and
LTCI, Télécom Paris, Institut Polytechnique de Paris, France}

\maketitle

\begin{abstract}
While existing makeup style transfer models perform an image synthesis whose results cannot be explicitly controlled, the ability to modify makeup color continuously is a desirable property for virtual try-on applications. We propose a new formulation for the makeup style transfer task, with the objective to learn a color controllable makeup style synthesis. 
We introduce CA-GAN, a generative model that learns to modify the color of specific objects (e.g. lips or eyes) in the image to an arbitrary target color while preserving background. 
Since color labels are rare and costly to acquire, our method leverages
weakly supervised learning for conditional GANs. This enables to learn a controllable synthesis of complex objects, and only requires a weak proxy of the image attribute that we desire to modify.
Finally, we present for the first time a quantitative analysis of makeup style transfer and color control performance.


\end{abstract}

\section{Introduction}


\begin{figure}[!] 
\centering
\includegraphics[width=8.7cm]{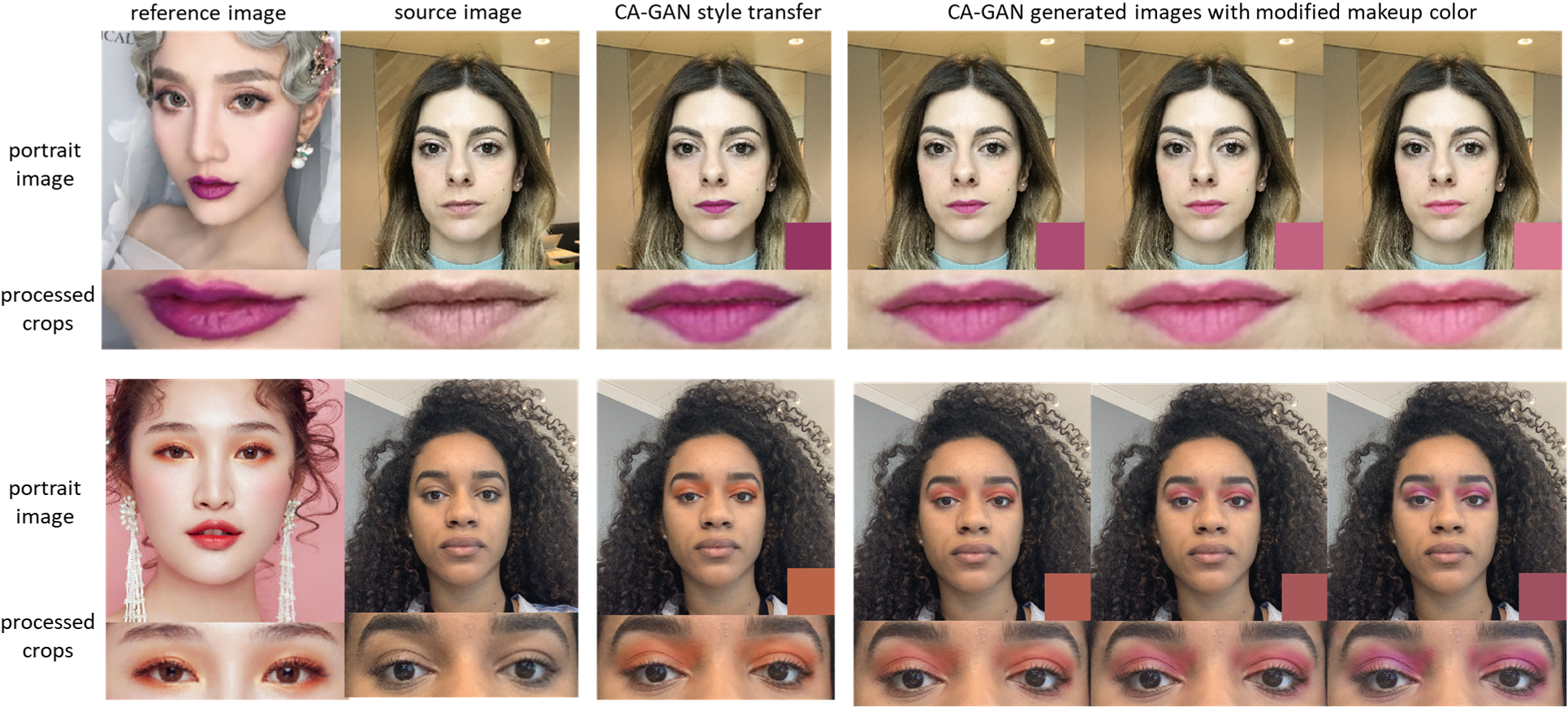}
\caption{Our CA-GAN model performs a color controllable makeup style transfer. The makeup color is explicitly estimated from the reference image and passed to the generator. Represented at the bottom right corner of each image, the makeup color can be modified to explore makeup style and reach the desired result.
}
\label{fig:fullface_shades}
\end{figure}

The development of online cosmetic purchase has led to a growing interest in makeup virtual try-on technologies. Based on image filtering \cite{sokalhigh} or physical modeling of skin and makeup optical properties \cite{li2015simulating}, makeup can be virtually applied to a source portrait image. 
Furthermore, thanks to the development of real-time facial landmark tracking \cite{kazemi2014one}, consumers can now try new cosmetics directly from their smartphone using Augmented Reality (AR) applications~\cite{modiface,perfectcorp}. However, conventional makeup rendering models often fail to take into account complex appearance
effects such as specular highlights.
In addition, makeup is applied on face pixels according to an estimated segmentation mask. This can lead to large failures for images with extreme facial poses, which are common when trying cosmetics such as lipsticks. 


More recently, the development of style transfer and image-to-image translation based on neural networks has led to new advances in the domain of makeup synthesis. 
The task of makeup style transfer, which consists in extracting makeup style from a reference portrait image, and applying it to the target image of a different person, has been widely studied 
\cite{chang2018pairedcyclegan,chen2019beautyglow,guo2009digital,li2018beautygan,liu2016makeup,tong2007example}. 
In contrast to standard augmented reality, such methods can implicitly model and transfer more complex makeup in a realist manner. Yet, makeup style transfer models suffer from a lack of control as the generated makeup style cannot be modified by the user to explore different cosmetic shades. 
Consequently, the obtained rendering cannot be transformed to simulate another close makeup shade or a target cosmetic product. Furthermore, the ability to try various shades is an indispensable characteristic expected by consumers in virtual try-on applications. 

In this paper, we propose to develop a makeup style transfer method in which the user can have fine control over the color of the synthesized makeup. 
Our main contributions can be summarized as follows:

\begin{itemize}[leftmargin=1cm, itemsep=1mm]
\item We propose CA-GAN, a color aware conditional gan that can modify the color of specific objects in the image to an arbitrary target color. This model is based on the use of a color regression loss combined with a novel background consistency loss that preserves the color attributes of non-targeted objects.  
\item To remove the need for costly color labeled data, we introduce weakly supervised learning for GAN based controllable synthesis.
This method enables to learn a controllable synthesis of complex objects, and only requires a weak proxy of the image attribute that we desire to modify.
\item We share a novel makeup dataset, the \textit{social media} dataset \footnote{available upon demand at \textit{contact.ia@rd.loreal.com}} of 9K images, with largely increased variability in skin tones, facial poses, and makeup color. 
\item For the first time, we introduce a quantitative analysis of color accuracy and makeup style transfer performance for lipsticks cosmetics using ground-truth images and demonstrate that our model outperforms state of the art. 
\end{itemize}

\section{Related Work}
In this section, we review related work on image synthesis and makeup style transfer. We first review GAN based methods for image-to-image translation that is the starting point of our approach. Then, we describe recent advances in controllable image-to-image synthesis using GANs. Finally, we present existing popular approaches for makeup style transfer.

\paragraph{GANs for Image-to-Image Translation.}

GAN based methods are at the origin of a large variety of recent success in image synthesis and image-to-image-translation tasks.
The idea of adversarial training of a discriminator and a generator model was first introduced in \cite{goodfellow2014generative}. Then, this method was extended in \cite{isola2017image} to image-to-image translation with conditional GANs. However, this method requires the use of pixel aligned image pairs for training, which is rare in practice. To overcome this limitation,  
the cycle consistency loss was introduced in~\cite{Zhu_2017_ICCV}, allowing to train GAN for image-to-image translation from unpaired images.
The use of GAN for solving image-to-image translation problems has later been extended to many different applications such as image completion \cite{iizuka2017globally}, super-resolution \cite{ledig2017photo} or video frame interpolation~\cite{jiang2018super}. 

\paragraph{Controllable Image Synthesis with GANs.}

In the field of GANs, efforts have recently been made to develop methods
that can control one or more \textit{attributes} of the generated images. A first research direction gathers works that attempt to implicitly control the model outputs in an unsupervised manner, through operations in the latent space. Among them, InfoGAN~\cite{chen2016infogan} aims to learn interpretable representations in the latent space based on information regularization. Furthermore, StyleGAN~\cite{karras2019style} is an architecture that leverages AdaIn layers~\cite{huang2017arbitrary} to implicitly diversify and control the style of generated images at different scales. More recently, an unsupervised method was proposed in \cite{voynov2020unsupervised} to identify directions in a GAN model latent space that are semantically meaningful. 
However, while these methods introduce a meaningful modification of the generated images, they have no control over which attributes are edited. Hence, the meaning of each modified attribute is described \textit{a posteriori} by the researchers while observing empirically the induced modification (``zoom", ``orientation", ``gender", etc.)
On the other hand, other studies attempt to provide explicit control of the generated images through supervised methods that leverage image labels. For instance, the method in \cite{lample2017fader} achieves continuous control along a specific class attribute by using adversarial training in the latent space. Besides, the StarGAN architecture in \cite{choi2018stargan} extends image-to-image translation to multiple class domains. This provides control over multiple attributes simultaneously, each being encoded as a discrete class. Later, in \cite{choi2020stargan}, inspired by the success of StyleGAN \cite{karras2019style}, the StarGAN architecture was improved using a style vector to enforce diversity of generated images within each target classes domain. However, it cannot be directly extended to continuous attributes.
While some studies attempt to modify color attributes, they only provide control on discrete color categories \cite{choi2018stargan} (e.g. ``blond hair", ``dark hair"),  or on the intensity of a discrete color class~\cite{lample2017fader}. 
Other synthesis methods are based on sketch conditions, that might contain color information as in~\cite{portenier2018faceshop}. However, in practice, such conditions can be complex for non-artist users and are not adapted to consumer-level applications. To the best of our knowledge, there is no existing method that enables high-level color control to an arbitrary shade in the continuous color space.

\paragraph{Makeup style transfer.}


The task of makeup style transfer has drawn interest throughout the evolution of computer vision methods. 
Traditional image processing methods such as image analogy \cite{hertzmann2001image} were first applied to this problem in~\cite{tong2007example}. Other early methods, such as in~\cite{guo2009digital}, propose to decompose an image into face structure, skin and color layers and transfer information between corresponding layers of different images. 
Later, neural networks based style transfer~\cite{gatys2016image} were used for makeup images~\cite{liu2016makeup}. However, such a method requires aligned faces and similar skin tones in source and target images.
Inspired by recent successes in  GANs, makeup style transfer was formulated in~\cite{chang2018pairedcyclegan} as an asymmetric domain translation problem. The authors of this work jointly trained a makeup transfer and makeup removal network using a conditional GAN approach.
In a later work, BeautyGAN~\cite{li2018beautygan} improved this GAN based approach by introducing a makeup instance-level transfer in addition to the makeup domain transfer. This is ensured through makeup segmentation and histogram matching between the source and the reference image. Furthermore, in~\cite{gu2019ladn}, makeup style transfer models were extended from processing local lips and eyes patches to the entire region of the face by using multiple overlapping discriminators. Such an improvement allows accurately transferring extreme makeup styles.

However, existing methods suffer from several limitations. First, the makeup extracted from the reference image is represented implicitly. It is therefore impossible to associate the synthesized makeup style with an existing cosmetic product that could be recommended to obtain that look. Furthermore, once the makeup style has been transferred, the generated image cannot be modified to explore other makeup shades. Prior studies \cite{chen2019beautyglow,liu2016makeup,zhang2019disentangled} attempted to propose makeup style transfer methods that are controllable, but only in terms of transfer intensity.

\section{Problem Formulation}

We propose a new formulation for the makeup style transfer problem, where the objective is to learn a color controllable makeup style synthesis. Hence, we propose to train a generator $G$ to generate a makeup style of an arbitrary target color $c$ from source image $x$. Furthermore, in order to perform makeup style transfer from reference image $y$ to source image $x$ we also need to train a discriminator $D_{color}$ to estimate the makeup color $c^y$ from $y$. Equation \ref{eq_objective} describes the objective of color controllable makeup style transfer, where $c^y$ belongs to a continuous three-dimensional color space:
\begin{equation}\label{eq_objective}
G(x, c^y) = G(x, D_{color}(y))
\end{equation}
With this new objective, the makeup color is transferred from the reference to the source image, and at the same time, explicitly controlled to reach the desired result. Furthermore, the estimated makeup color can be used to compute a correspondence with existing cosmetics products that can be recommended.
In contrast to other studies, we do not decompose between before and after makeup image domains. Indeed, in practice consumers desire to virtually try new shades without removing their current makeup. For this reason, it is desirable to train a model that can generate makeup style from portrait images with or without makeup.
Finally, we desire to train our model from unlabeled unpaired images. Indeed, while massively available, makeup images are rarely qualified with labels on which cosmetics were used. Furthermore, there is currently no large database containing image pairs before and after makeup using the same makeup style, and collecting one would be particularly costly.

\section{CA-GAN: Color Aware GAN}

To solve this problem, we introduce the novel CA-GAN architecture, a color aware generative adversarial network that learns to modify the color of specific objects to an arbitrary target color. Our proposed model is not specific to makeup images and could be trained on any object category that can be described by a single color. Furthermore, the CA-GAN model does not require images with color labels since it can be trained in a weakly supervised manner. 
While the architecture of our model is close to existing popular methods, we introduce new losses for both generator and discriminator that are critical for accurate color control (see Section~\ref{sec:loss}).

\subsection{Weakly supervised color features}

\begin{figure}[t]
\centering
\includegraphics[width=13cm]{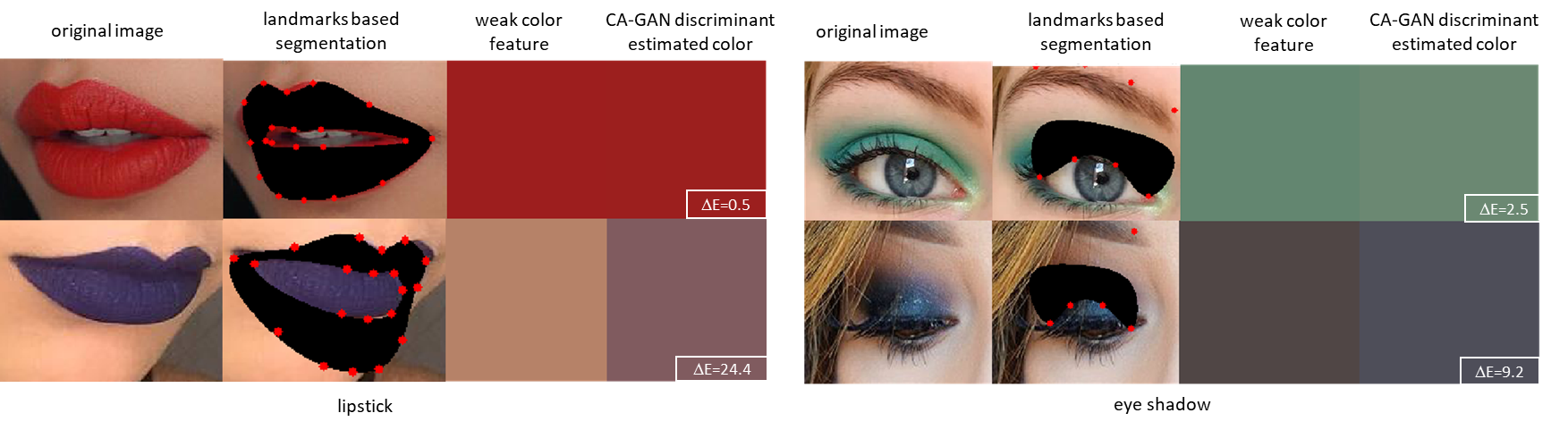}
\caption{Example on test images of $C_m(x)$ the weak makeup extractor versus $D_{color}(x)$ our learned color discriminator module. Estimated facial landmarks are represented as red dots. While the two models agree on many images (top row), our learnt model seems superior in case of disagreement (bottom row). }
\label{fig:makeup_stats}
\end{figure}

 Since our objective is to learn to modify the color of an object in the image to an arbitrary color, we need color values to support the training of our generator.
However, most available datasets do not contain labels on objects color. Furthermore, labeling the apparent color value of an object in an image is a tedious task that is highly subjective. On the other hand, GAN based models require a large amount of data to be trained. To overcome this difficulty, we introduce a method to train our model in a weakly supervised manner. 
 Instead of using manually annotated color labels, we propose to use a weak proxy for the target object color attributes that can be obtained without supervision. In particular, in the case of makeup, we build on the assumption that makeup is generally localized on specific regions of the face, which can be approximately estimated for each image using traditional face processing methods. We denote by $C_m(x)$ our weak makeup colors feature extractor, illustrated in Figure~\ref{fig:makeup_stats}. This weak estimator consists in first estimating the position of facial landmarks using the popular {\em dlib} library \cite{king2009dlib} and then computing the median pixel in a fixed region defined from landmarks position, for lips and eye shadow. Similarly, we also use $C_s(x)$, a weak skin color model to compute the skin color in each image, using the inverse makeup segmentation mask. Skin color will be used to ensure background color consistency when processing local crops. 
 
This color feature extractor is \textit{weak} in the sense that it produces a noisy estimate of a makeup color. The landmarks estimation often fails for complex poses, and the median color estimation does not take into account shading effects nor occlusion, as illustrated in Figure~\ref{fig:makeup_stats}. Furthermore, the spatial information on which $C_m(x)$ relies only captures a simplified information of the makeup style, in particular for eye makeup.
For this reason, we avoid to use $C_m(x)$ to directly control the generator output, and instead use it as a weak supervisor for $D_{color}(x)$ learned color discriminant module. By leveraging the noisy signal of $C_m(x)$ over a large amount of data, $D_{color}(x)$ learns a better representation for the attribute of interest, and outperforms $C_m(x)$ as discussed in Section 5.3. 

\subsection{CA-GAN architecture}

\begin{figure}[t!]
\centering
\includegraphics[width=7.5cm]{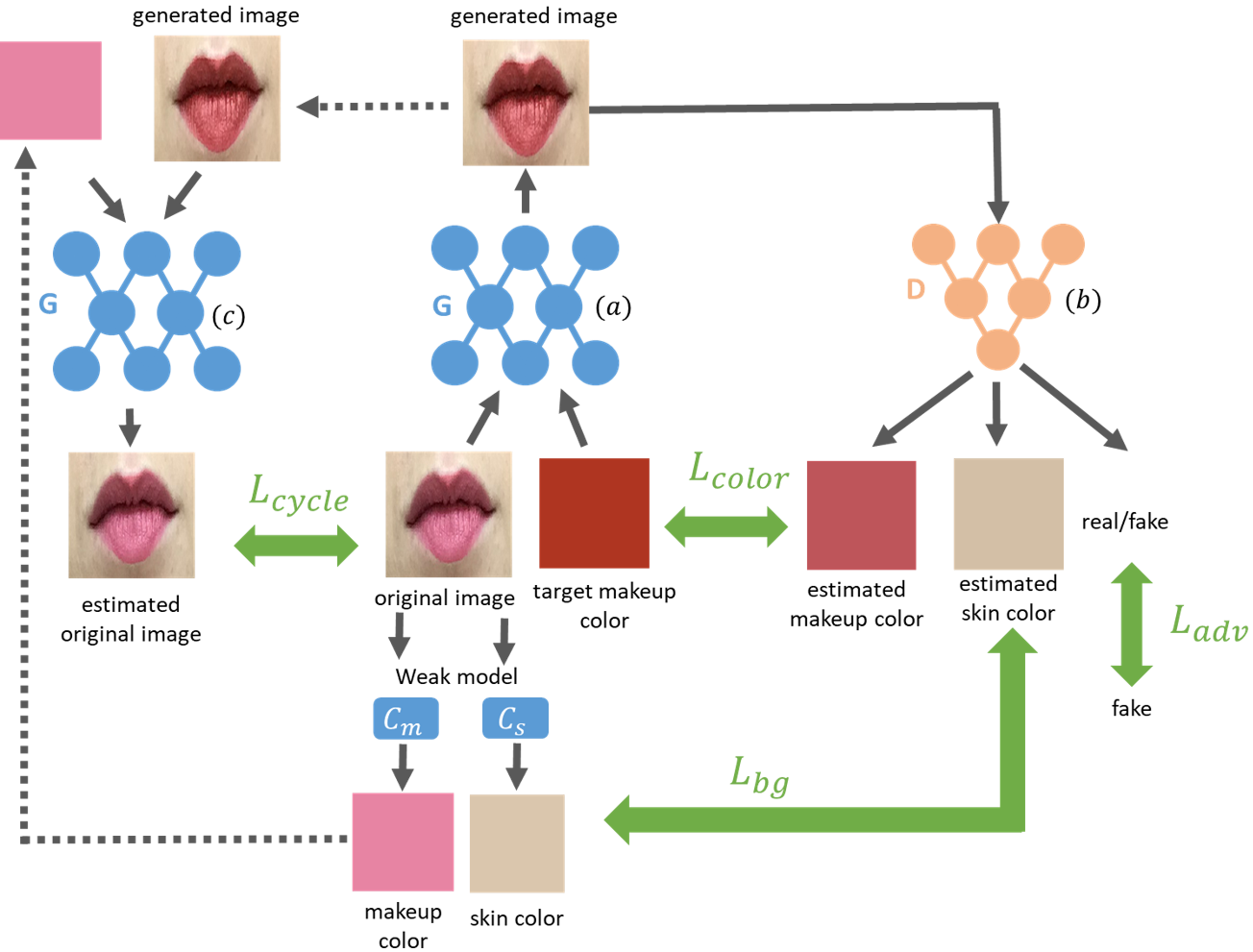}
\caption{The training procedure of our CA-GAN model. First (a) the generator $G$ estimates an image from a source image and a target makeup color. Secondly (b) the discriminator $D$ estimates the makeup color, skin color and a real/fake classification from the generated image, used to compute the color regression loss $L_{color}$, background consistency loss $L_{bg}$ and adversarial loss $L_{adv}$, respectively. Thirdly (c), the source image is reconstructed from the generated one using the makeup color as target. The reconstruction is used to compute the cycle consistency loss $L_{cycle}$.}
\label{fig:model_pipeline}
\end{figure}

Our CA-GAN model consists of two different networks, a generator and a discriminator, that are jointly trained. To achieve a higher resolution in the generated images, the model only processes crops of the region of interest. Besides, we train two independent CA-GAN models to process lips and eyes images.

\subsubsection{Generator}
Our generator takes as input a source image together with a target color and outputs an estimated image. Its architecture is described in detail in Table 1 of the supplementary material. 
As in StarGAN \cite{choi2018stargan} the input condition is concatenated as an additional channel of the source image. The residual blocks that we use consist of two convolutional layers with $4\times 4$ kernels and a skip connection. Similarly to \cite{chang2018pairedcyclegan}, the generator outputs a pixel difference that is added to the source image in order to obtain the generated image.

\subsubsection{Discriminator}
As described in Table 1 of the supplementary material,
our discriminator network is a fully convolutional neural network, similar to PatchGAN~\cite{isola2017image}, with multi-task output branches.  The discriminator network simultaneously estimates makeup color, skin color, and classifies the image as real or fake, as illustrated in Figure \ref{fig:model_pipeline}.

\subsection{CA-GAN objective function}
\label{sec:loss}

In this section, we introduce the loss functions that are the key components of our novel CA-GAN model. 
The training procedure is summarized in Figure \ref{fig:model_pipeline}.
 
 \subsubsection{Color regression loss} 
 
The color regression loss ensures that the makeup color in the generated image is close to the target color condition passed to the generator. 
During training, for each image $x_i$ among the $n$ training examples, a target color $c_i$ is randomly sampled at each epoch among existing colors in the training set.
The color regression loss computes a color distance between a target color $c_i$ and $D_{color}(G(x_i,c_i))$, the color of the generated image as estimated by the makeup color branch of the discriminator.
As a color regression loss, we propose to use $mse-lab$ , the mean squared error in the CIE $L^*a^*b^*$ space.
Introduced for neural networks in \cite{kips2019skin}, the $mse-lab$ loss inherits from the perceptual properties of the color distance \textit{CIE $\Delta$E* 1976} \cite{mclaren1976xiii} which is key for color estimation problems. The color regression loss  is described in Equations \ref{eq_color_d} and \ref{eq_color_g} for the discriminator and the generator respectively, where $D_{color}$ is the makeup color regression output of the discriminator, and $c_i^{x_i} = C_m(x_i)$ the color label for image $x_i$ obtained using our weak model:
\begin{equation}\label{eq_color_d}
 L_{color}^D = \frac{1}{n}  \sum_{i=1}^n  \left\lVert c_i^{x_i} - D_{color}(x_i) \right\rVert ^2
\end{equation}
\begin{equation}\label{eq_color_g}
 L_{color}^G = \frac{1}{n}  \sum_{i=1}^n \left\lVert c_i - D_{color}(G(x_i,c_i)) \right\rVert ^2
\end{equation}
 
\subsubsection{Adversarial loss}
As in any GAN problem, we use an adversarial loss whose objective is to make generated images indistinguishable from real images. In particular, we use the Wasserstein GAN loss \cite{arjovsky2017wasserstein} and more specifically the one from~\cite{gulrajani2017improved} with gradient penalty. Our used adversarial loss is described in Equation \ref{eq_adv_d} for the discriminator and Equation \ref{eq_adv_g} for the generator, where $D_{proba}$ is the realism classification output of the discriminator and ${\lambda}_{gp} \: gp(D)$ the weighted gradient penalty term computed on $D$:
\begin{equation}\label{eq_adv_d}
L_{adv}^D = \frac{1}{n} \sum_{i=1}^n  D_{proba}(G(x_i, c_i)) - \frac{1}{n} \sum_{i=1}^n  D_{proba}(x_i) + {\lambda}_{gp} \: gp(D)
\end{equation}
\begin{equation}\label{eq_adv_g}
L_{adv}^G = - \frac{1}{n} \sum_{i=1}^n  D_{proba}(G(x_i, c_i))
\end{equation}

\subsubsection{Cycle consistency loss} 
Since we are learning image-to-image translation from unpaired images, we need an additional loss to ensure that we will not modify undesired content in the source image. Consequently, we employ a cycle consistency loss described in Equation \ref{eq_cycle}, where we compute a perceptual distance between $x_i$ and its reconstruction $ \hat{x_i} = G(G(x_i,c_i), c_i^{x_i}))$. As a perceptual distance, we choose $MSSIM$, the multiscale structural similarity loss introduced by \cite{wang2003multiscale}, leading to:
\begin{equation}\label{eq_cycle}
L_{cycle} = 1 - MSSIM(x_i, \:\hat{x_i})
\end{equation}

\subsubsection{Background consistency loss} 
Since the generator is only processing local crops of the image, we need to ensure that the background color stays consistent with the rest of the image. Besides, if the background color is modified by the generator, the adversarial loss and the cycle consistency loss will not be able to penalize this change as it might lead to a realistic image and modify color in the same direction as the target color. Thus, we propose a background consistency loss that penalizes the color modification of the background. In the case of makeup color, the background color is represented by the skin color on the source image. 
Equations \ref{eq_bkg_d} and \ref{eq_bkg_g} describe background consistency for the discriminator and the generator, respectively, where $D_{bg}$ is the background color estimation output of the discriminator and $b_i^{x_i} = C_s(x_i)$ the extracted background color of the image $x_i$:
\begin{equation}\label{eq_bkg_d}
L_{bg}^D = \frac{1}{n}  \sum_{i=1}^n \left\lVert b_i^{x_i} - D_{bg}(x_i) \right\rVert ^2
\end{equation}
\begin{equation}\label{eq_bkg_g}
 L_{bg}^G =  \frac{1}{n} \sum_{i=1}^n \left\lVert D_{bg}(x_i) - D_{bg}(G(x_i,c_i))\right\rVert ^2
 \end{equation}

\subsubsection{Total objective functions}
Finally, to combine all the loss functions, we propose to use weighting factors for each loss of the generator. Indeed, some factors such as the cycle consistency loss and the reconstruction loss must be balanced as they penalize opposite transformations. Equations \ref{eq_total_d} and \ref{eq_total_g} describe the total objective functions of the discriminator and the generator, where $\lambda_{color}$, $\lambda_{bg}$ and $\lambda_{cycle}$ are weighting factors for each generator loss that are set experimentally:
\begin{equation}\label{eq_total_d}
L_D = L_{adv}^D + L_{color}^D + L_{bg}^D
\end{equation}
\begin{equation}\label{eq_total_g}
L_G = L_{adv}^G + \lambda_{color} \: L_{color}^G + \lambda_{bg} \: L_{bg}^G + \lambda_{cycle} \: L_{cycle}
\end{equation}

\section{Experiments}

\begin{figure}[t!]
\centering
\includegraphics[width=12.7cm]{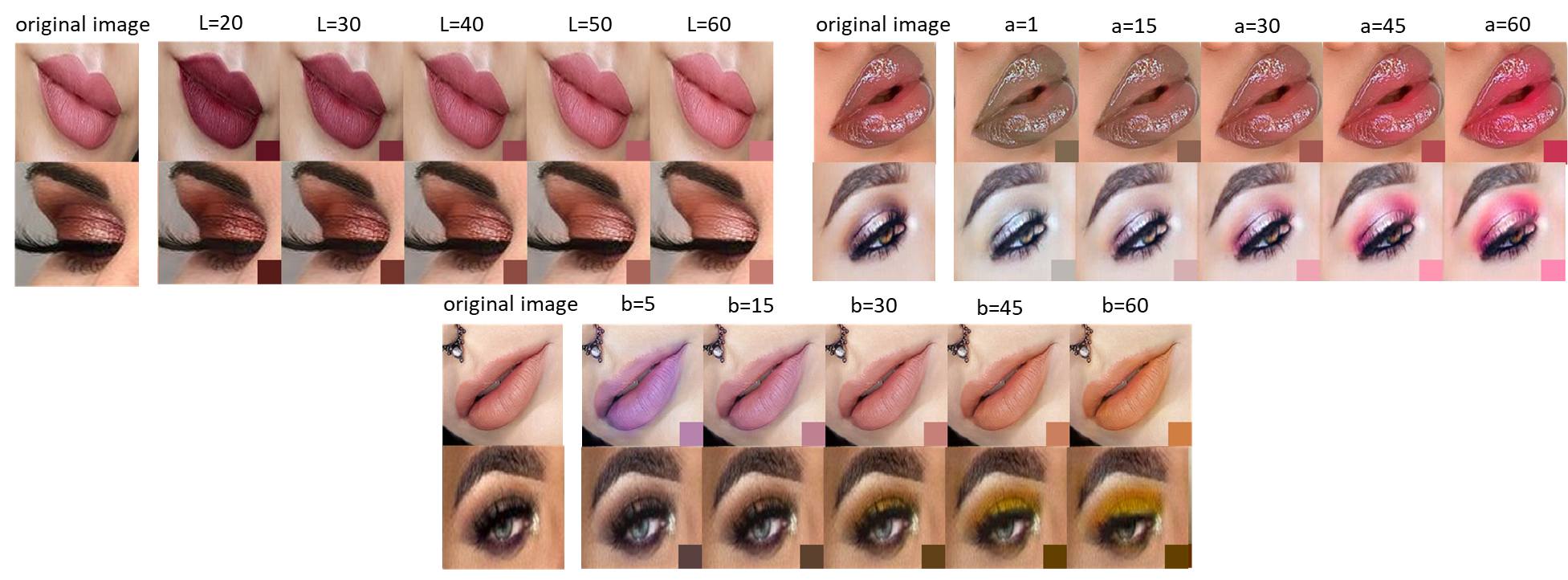}
\caption{Modification of makeup color along each dimension of the $CIE L^*a^*b^*$ color space, using images from our social media dataset. The color patch on the bottom-right of each image illustrates the target color passed to the model. Our approach generalizes to lips and eyes images with various makeup textures and facial poses.}
\label{fig:closeup_editing}
\end{figure}

\subsection{Data}

Since our model does not require images before and after makeup to be trained, we are not restricted to the conventional makeup style transfer datasets such as the MT dataset \cite{li2018beautygan}. Instead, we collected a database of 5000 social media images from makeup influencers. Compared to MT, this dataset contains a larger variety of skin tones, facial poses, and makeup color, with 1591 shades of 294 different cosmetics products. Since these images are unpaired and unlabeled, they are used to train our model using our proposed weakly supervised approach. We will refer to this database as the social media dataset.
Furthermore, for model evaluation purposes, we collected a more controlled database focusing on the lipstick category. Therefore, we gathered images of 100 panelists with a range of 80 different lipsticks with various shades and finish. For each panelist, we collected images without makeup, and with three different lipsticks drawn from the 80 possible shades. This dataset will be referred to as the lipstick dataset.

\subsection{Implementation}

Our CA-GAN model is implemented using the Tensorflow \cite{abadi2016tensorflow} deep learning framework. The generator and discriminator are jointly trained on 90 percent of our social media dataset, with lips and eyes crops of size 128 by 128 pixels. The weighting factors of the generator loss are set to $\lambda_{gp}=10$,  $\lambda_{color}=10$, $\lambda_{bkg}=5$, $\lambda_{cycle}=200$. We train our model over 200 epochs using the adam optimizer \cite{kingma2014adam}  with a learning rate of $10^{-3}$ for the discriminator and $3.10^{-3}$ for the generator.
Finally, we train separated CA-GAN models for processing lips and eyes images, as well as a joint model trained on both categories. As illustrated in Section \ref{quant_eval} and the supplementary material, separated models slightly overperform the joint model, and are thus used for the image results presented in this study.

\subsection{Qualitative evaluation}

\subsubsection{Color controllable makeup synthesis}

\begin{figure}[t!]
\centering
\includegraphics[width=7cm]{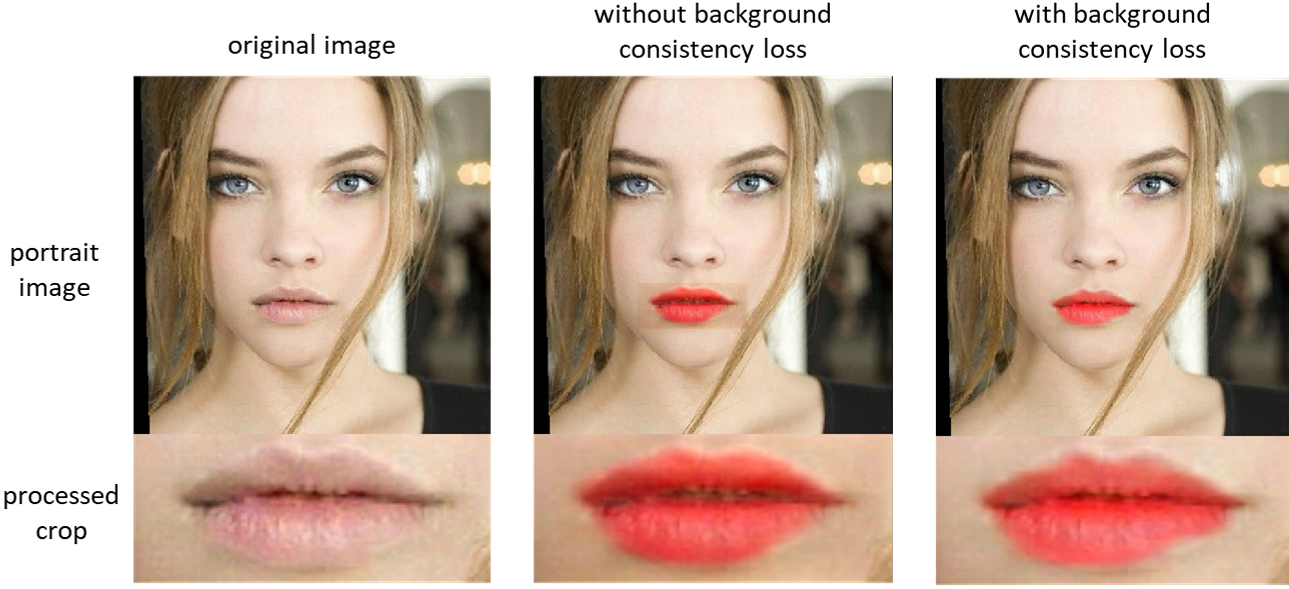}
\caption{Our background consistency loss improves the preservation of the skin color in the modified image, which is essential at the portrait scale.}
\label{fig:skin_consistancy}
\end{figure}

First, we use images from our social media dataset that are unseen during training and modify their makeup color independently in each dimension of the $CIE \: L^*a^*b^*$ color space. This experiment intends to illustrate the performance of our model with complex poses and makeup textures, and the results are displayed in Figure \ref{fig:closeup_editing}. 
In addition, we generate portrait images typically encountered in augmented reality tasks using our lipstick dataset, visible in Figure \ref{fig:fullface_shades}. 
For both experiments, it can be observed that the synthesized images reach well the target color while preserving their realistic appearance.
Our approach generalizes well for both lips and eye images, with various skin colors, makeup colors, and textures. In particular, for images of eyes without makeup, eye shadow seemed to be synthesized on an average position around the eye, as visible in Figure~\ref{fig:fullface_shades}. 
Furthermore, our model implicitly learns to only modify the makeup color attributes, preserving other dimensions such as shine or eye color, as it can be observed in Figure \ref{fig:closeup_editing}. Such results are usually obtained through complex image filtering techniques and would need a specific treatment depending on each object category. 
Additional generated images and videos \footnote{also accessible at \url{https://robinkips.github.io/CA-GAN/} }
are presented in the supplementary material, illustrating performance on various skin tones and illuminants.

\subsubsection{Skin color preservation}

Even though our model only processes a local crop of the image, the color of skin pixels is preserved, and the crop modification is not easily perceivable at the portrait scale, as seen in Figure \ref{fig:fullface_shades}. For this reason, we do not need to use Poisson blending to insert the processed crop in the final image as used in \cite{chang2018pairedcyclegan,chen2019beautyglow}, which speeds up computations and avoids using a segmentation of the lips or eyes region.
As an ablation study, we train a CA-GAN model without using the proposed background consistency loss. As observed in Figure~\ref{fig:skin_consistancy}, skin pixels are also modified in the generated image. Even though these changes might look realistic at the patch level and thus are not penalized by the adversarial loss, they are not acceptable at the portrait image level. Using our background consistency loss however, skin color modification is penalized by the discriminator, which leads to significantly improved results. 

\begin{figure}[t]
\centering
\includegraphics[width=8cm]{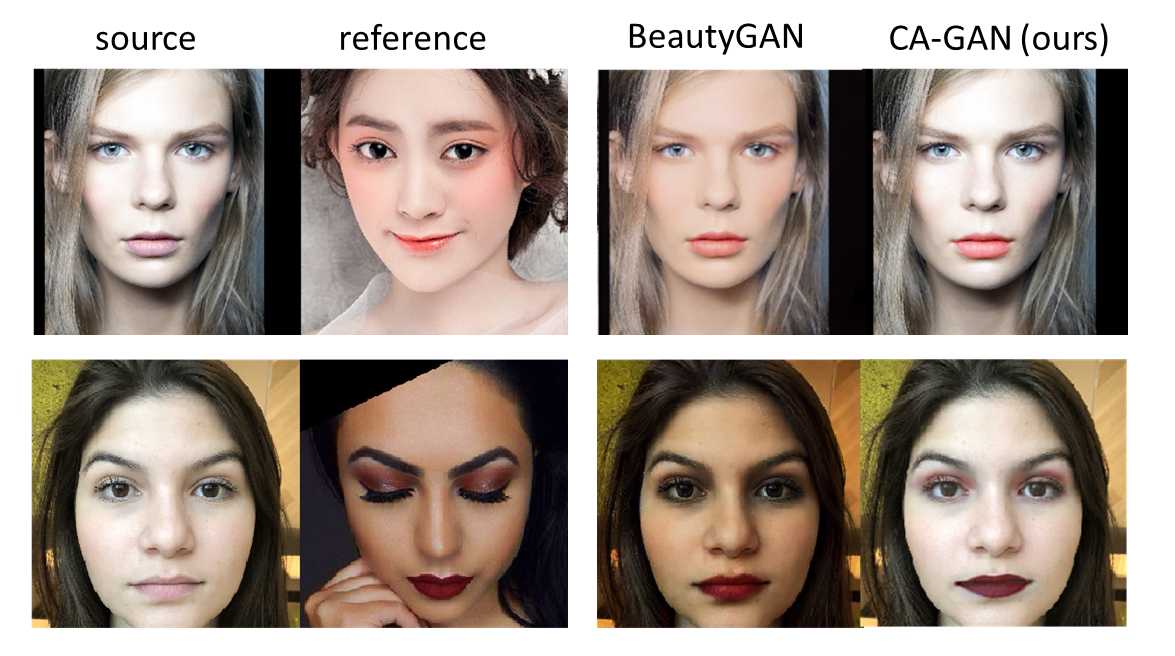}
\caption{Our model shows makeup style transfer performances that are equivalent to state of the art models, while obtaining better preservation of the skin color of the source subject. More results are presented in the supplementary material.}
\label{fig:style_transfert_quali}
\end{figure}

\subsubsection{Makeup style transfer}

We use our lipstick dataset as typical source images, and perform makeup style transfer from reference images drawn from the MT dataset, as illustrated in Figure~\ref{fig:fullface_shades}. In addition to obtaining a realistic generated image, the makeup style can be edited to explore other makeup styles in a continuous color space.
Furthermore, our model also estimates the makeup color which can be used to recommend existing cosmetics that can be used in practice to achieve a similar result. 
Moreover, we compared our results on the style transfer tasks against other popular models for which the code is available. To perform style transfer with our CA-GAN model, we estimate the makeup color in the reference image using the color regression branch of the discriminator, and generate a synthetic makeup image using the generator. The obtained results can be observed in Figure \ref{fig:style_transfert_quali}. We compared our model against BeautyGAN \cite{li2018beautygan} which is a state of the art method for conventional makeup style transfer.
Our model transfers makeup color with equivalent performance. Furthermore, while BeautyGAN tends to transfer the skin color together with the makeup style, our model obtains better preservation of the original skin tone of the source subject, which is a desirable property for virtual try-on applications.

\subsubsection{Weak vs learned color estimator}

\begin{figure}[t]
\centering
\includegraphics[height=3.0cm]{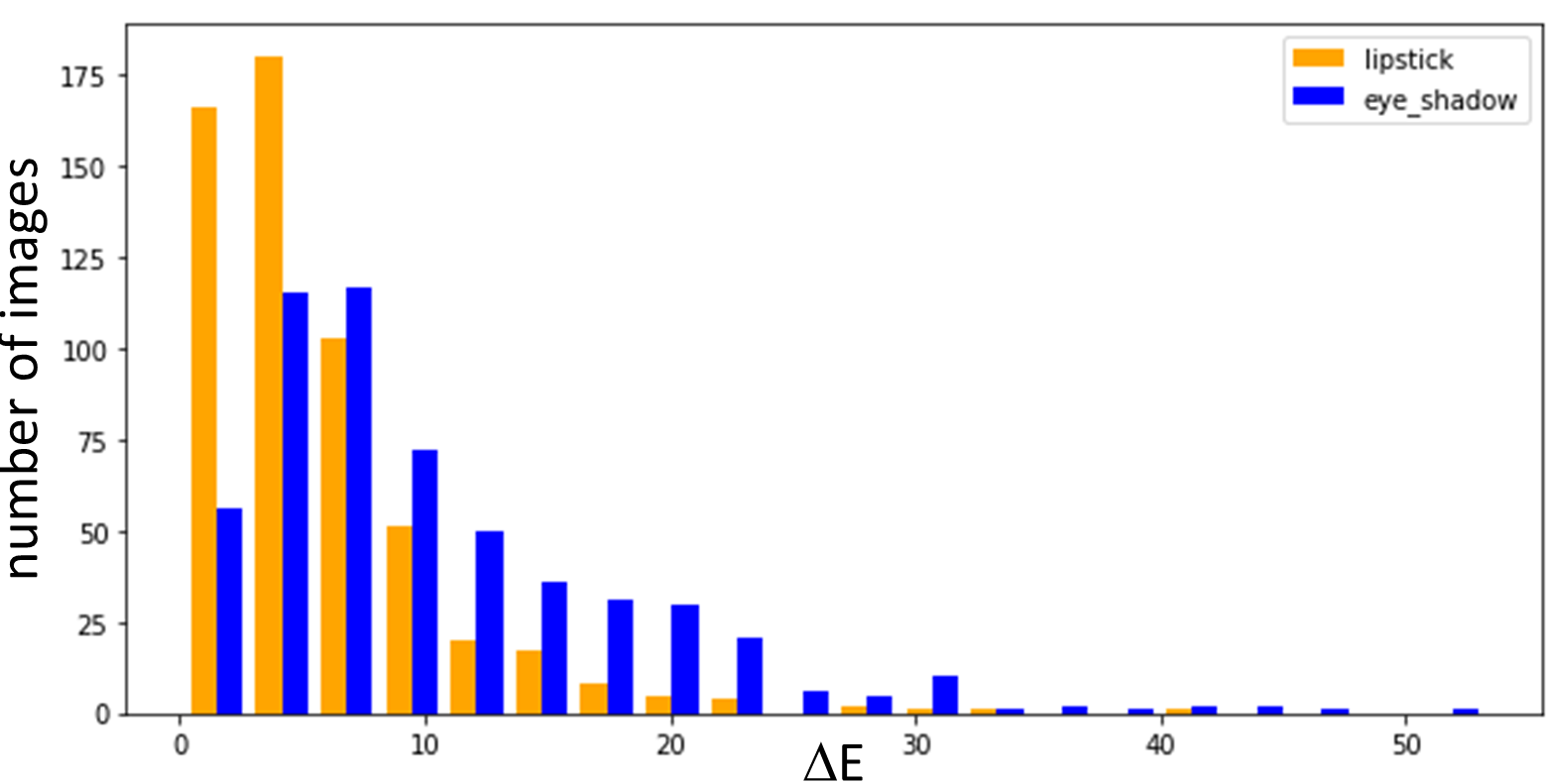}
\caption{Color difference between weak color features and learnt discriminant. Large differences between the two models are generally due to failure of the weak feature extractor.}
\label{fig:hist_weak_vs_learned}
\end{figure}

While the weak color estimator $C_m(x)$ used for weak supervision is fixed, the learnt color extractor in the discriminant $D_{color}(x)$ leverages a large dataset. Hence, even if $C_m(x)$ has high variance and largely fails for some images, $D_{color}(x)$ learns a more robust color estimator. To illustrate this idea, we computed on test images the color difference between estimates of the weak model and the corresponding learned discriminant, as illustrated in Figure~\ref{fig:hist_weak_vs_learned}. 
Even if the two models agree for most images, large differences occur in some cases. In practice, we found that in most large difference cases, the weak estimator was failing due to poor facial landmark localization, occlusion, or complex appearance with shading and specularities (see Figure~\ref{fig:makeup_stats} and supplementary material). The difference is even larger for the eye shadow region in which appearance is more complex due to hair and eyelash occlusion. This reinforces the interest of weakly supervised learning for GAN based model, since improved color estimation will improve the generator control and in turn the style transfer accuracy. 

\subsection{Quantitative evaluation \label{quant_eval}}

 In this section, we focus on the evaluation of the model on lips images. Indeed, while there is no existing approach for eye makeup segmentation, that might be on a larger region than the eyelid, it is possible to segment the lips makeup region for our experiments using face parsing models.

\subsubsection{Color accuracy evaluation}
First, we evaluate the ability of our CA-GAN model to generate makeup images that are close to the chosen color target. For this experiment, illustrated in the supplementary material, we use the 500 test images from our social media dataset. First, we choose a set of 50 representative lipstick shades by computing the centroids of a k-means clustering of the lipstick colors in our training data. Then, for each test sample, we generate an image with each representative lipstick color as target. Finally, using a lips segmentation algorithm we estimate the median color of lips to compute a color distance to the model target. We also estimate the difference between the color of the skin before and after image synthesis to control its preservation. The results of these experiments are reported in Table \ref{tab:results_color}.
The ablation study confirms that the use of the $lab-mse$ for color loss largely increases the color accuracy of our model. Furthermore, our novel background consistency loss helps the generator to disentangle skin and lips color, which leads to significantly improved lipstick color accuracy and skin color preservation.

\begin{table*}[t]
 \centering
 \setlength{\tabcolsep}{6pt}
 \caption{The ablation study demonstrates that our color regression loss and background consistency loss significantly increase the makeup color synthesis accuracy and skin color preservation.}
 \scalebox{0.75}{
 \begin{tabular}{|c|c|c|c|c|c|}
 \hline
 Model & color loss & \thead{ background \\ consistency loss} & training images & \thead{lips color accuracy \\ ($\Delta$E mean)}  & \thead{ skin color preservation \\ ($\Delta$E mean) }\\ [0.5ex] 
 \hline\hline
  CA-GAN & rgb-mse & no & lips & 25.82 & 19.49 \\ 
 \hline
 CA-GAN & lab-mse & no & lips & 9.62  & 10.18 \\ 
 \hline
 CA-GAN & lab-mse & yes & lips & \textbf{6.80} & \textbf{6.05} \\ 
 \hline
  CA-GAN & lab-mse & yes & eyes and lips & 7.78 & 8.76 \\ 
 \hline
  \end{tabular}}
  \label{tab:results_color}
\end{table*}

\subsubsection{Style transfer performance evaluation}

\begin{figure}[!]
\centering
\includegraphics[height=3cm]{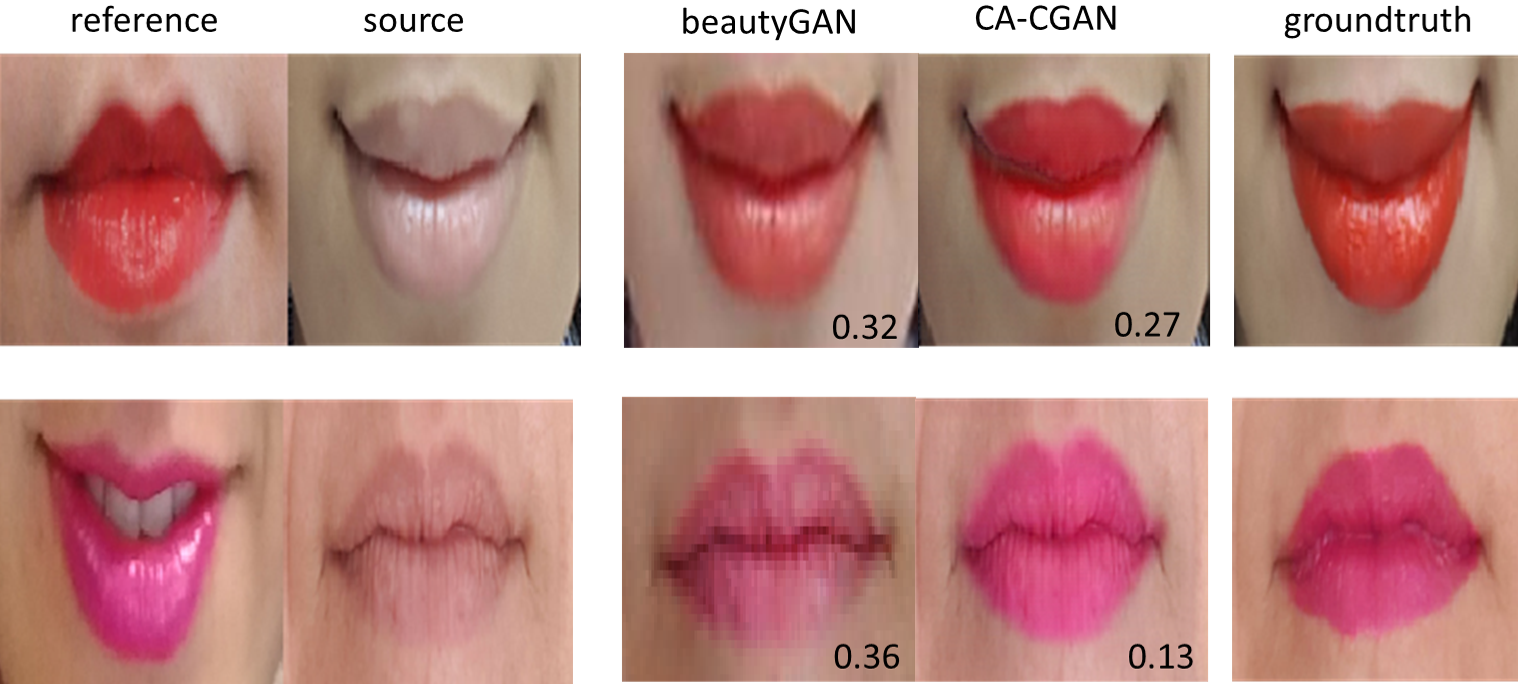}
\caption{The style transfer performance is evaluated using triplets of lips images. The makeup is extracted from the reference image and transferred to the source image of a different panelist. We use a ground-truth image of the source panelist with the same lipstick to compute a style transfer performance. The computed perceptual distance $1- MSSIM$ is given at the bottom right of each generated image.}
\label{fig:example_style_stransfert_quant}
\end{figure}

For the first time, we introduce a quantitative evaluation of model performance on the makeup style transfer task, as illustrated in Figure \ref{fig:example_style_stransfert_quant}. We use our collected lipstick dataset that contains images of multiple panelists wearing the same lipstick shade. Thus, it is possible to construct ground-truth triplets with a reference portrait, a source portrait, and the associated ground-truth image with the reference makeup. The style transfer accuracy is then computed using the MSSIM similarity \cite{wang2003multiscale} as a measure of a perceptual distance. Furthermore, to avoid lighting bias, we select the ground-truth among several images of the same panelist, using the most similar skin color compared to the source image. We perform this experiment on 300 image triplets with 100 different panelists and 80 different lipstick shades.
The results of this experiment are reported in Table \ref{tab:results_style_transfert}. The ablation study confirms that our color regression loss and background consistency loss significantly improve the style transfer performance. Furthermore, we observe that our model outperforms BeautyGAN by a significant margin. This is expected given the ability of our model to preserve the skin color in the source image.

\begin{table*}[!]
 \centering
 \setlength{\tabcolsep}{8pt}
 \caption{A quantitative evaluation of the style transfer performance using style transfer image triplets.}
 \scalebox{0.7}{
 \begin{tabular}{|c|c|c|c|c|c|c|}
 \hline
 Model & color loss &  \thead{background consistancy loss} & training images & L1 & 1 - MSSIM \\ [0.5ex]
 \hline\hline
 BeautyGAN \cite{li2018beautygan} & - & - & - & 0.124 & 0.371\\ 
 \hline
  CA-GAN & rgb-mse & no & lips & 0.231 & 0.698 \\ 
 \hline
 CA-GAN & lab-mse & no & lips & 0.097 & 0.313\\ 
 \hline
 CA-GAN & lab-mse & yes & lips & \textbf{0.085} & \textbf{0.283}\\ 
 \hline
  CA-GAN & lab-mse & yes & eyes and lips & 0.087 & 0.312\\ 
 \hline
  \end{tabular}}
  \label{tab:results_style_transfert}
\end{table*}




\section{Conclusion and Future Work}

In this paper, we introduced CA-GAN, a generative model that learns to modify the color of objects in an image to an arbitrary target color. This model is based on the combined use of a color regression loss with a novel background consistency loss that learns to preserve the color of non-target objects in the image. Furthermore, CA-GAN can be trained on unlabeled images using a weakly supervised approach based on a noisy proxy of the attribute of interest. 
Using this architecture on makeup images of eyes and lips we show that we can perform makeup synthesis and makeup style transfer that are controllable in a continuous color space.
For the first time, we introduce a quantitative analysis of makeup style transfer and color control performance. 
Our results show that our model can accurately modify makeup color, while outperforming conventional models such as \cite{li2018beautygan} in makeup style transfer realism. 
Since our CA-GAN model does not require labeled images, it could be directly applied to other object categories for which it is possible to compute pixel color statistics, such as hair, garments, cars, or animals. 

Finally, we emphasize some perspectives for future work. First, we represent eyes and lips makeup by three-dimensional color coordinates. However, extreme makeup can be composed of multiple different cosmetics, in particular for the eye shadow category. To achieve color control on multiple cosmetics simultaneously, our model should be extended with a spatial information condition in addition to our current color condition. However, 
while our model can currently be trained in a weakly supervised manner, using segmentation masks to carry the spatial information would require to have annotated images.
Moreover, the representation of cosmetics could also be completed using a shine representation. While the current model objective is to learn to modify color only,  without affecting the other image attributes such as shine and specularities,  using a shine score as an additional generator condition would make it possible to simulate mat and shine cosmetics with more accuracy.

%
%


\bibliographystyle{splncs04}
\bibliography{eccv2020submission}
\end{document}